\documentclass[sigconf]{acmart}
\AtBeginDocument{%
  \providecommand\BibTeX{{%
    \normalfont B\kern-0.5em{\scshape i\kern-0.25em b}\kern-0.8em\TeX}}}

\setcopyright{acmcopyright}
\copyrightyear{}
\acmYear{}

\acmBooktitle{}

\acmPrice{15.00}
\acmISBN{978-1-4503-XXXX-X/18/06}
\acmDOI{XXXXXXX.XXXXXXX}

\usepackage{multirow}
\usepackage{color}
\usepackage{svg}
\usepackage{graphicx}
\usepackage{subfigure}
\newcommand\wx[1]{{\color{black}#1}}

\graphicspath{ {figures/} }

\begin{document}

\title{
{Adaptive Semantic Consistency for Cross-domain Few-shot Classification} 
}

\author{Hengchu Lu$^1$, Yuanjie Shao$^2$, Xiang Wang$^1$, Changxin Gao$^1$}
\affiliation{%
  \institution{$^1$School of Artificial Intelligence and Automation, Huazhong University of Science and Technology}
  \institution{$^2$School of Electronic Information and Communications, Huazhong University of Science and Technology}
  \streetaddress{}
  \city{}
  \state{}
  \country{}
  }
\email{{u202015050, shaoyuanjie, wxiang, cgao}@hust.edu.cn}



\begin{abstract}
\wx{Cross-domain few-shot classification (CD-FSC) aims to identify novel target classes with a few samples, assuming that there exists a domain shift between source and target domains.}
\wx{Existing state-of-the-art practices typically pre-train on source domain and then finetune on the few-shot target data to yield  task-adaptive representations. 
}
\wx{
Despite promising progress, these methods are prone to overfitting the limited target distribution since data-scarcity and ignore the transferable knowledge learned in the source domain.
}
\wx{To alleviate this problem, we propose a simple plug-and-play \textbf{A}daptive \textbf{S}emantic \textbf{C}onsistency (\textbf{ASC}) framework, 
which improves cross-domain robustness by preserving source transfer capability during the finetuning stage. 
}
\wx{
Concretely, we reuse the source images in the pretraining phase and design an adaptive weight assignment strategy to highlight the samples similar to target domain, aiming to aggregate informative target-related knowledge from source domain.
}
\wx{
Subsequently, a semantic consistency regularization is applied to constrain the consistency between the semantic features of the source images output by the source model and target model.
}
\wx{
In this way, the proposed ASC enables explicit transfer of source domain knowledge to prevent the model from overfitting the target domain.
}
\wx{
Extensive experiments on multiple benchmarks demonstrate the effectiveness of the proposed ASC, and ASC provides consistent improvements over the baselines.
The source code is released at \href{https://github.com/luhc666/ASC-CDFSL}{https://github.com/luhc666/ASC-CDFSL}.
}

\end{abstract}





\maketitle
\begin{figure}[htbp]
\centering
\includegraphics[width=\linewidth]{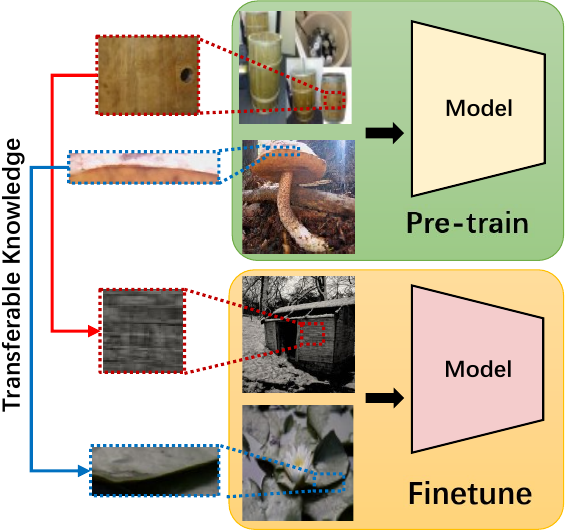}
\caption{An illustration of prior learned target-related knowledge. The source images used in pre-train phase contain similar semantic features as target images. The capability of extracting such transferable features should be preserved in finetuning phase.
}
\label{figure1}
\end{figure}
\section{Introduction}
\label{1}

\wx{
Few-shot learning~\cite{snell2017prototypical, vinyals2016matching, sung2018learning, finn2017model,chen2019closer,dhillon2019baseline, tian2020rethinking} attempts to classify unseen classes with limited labeled samples, which is a promising direction to alleviate the high manpower and time cost problems caused by massive data annotation.
}
\wx{
Existing few-shot methods mainly focus on optimizing a model on the base (source) dataset during the meta-training stage and quickly adapting the learned model to the unseen target classes with only a few samples in the meta-testing phase.
Despite impressive results, these practices typically assume that the base and target datasets share the same domain distribution.
}
\wx{
However, in real-world application scenarios, there may exists domain shifts between the source and target datasets, and thus cross-domain few-shot classification (CD-FSC) is introduced and has attracted increasing attention in recent years~\cite{chen2019closer, guo2020broader}.}

\wx{ 
Existing CD-FSC methods usually adopt the two-stage optimization manner, which first pretrains a model on the source domain and then leverages the few-shot target samples to further finetune the learned model.
}
\wx{
For instance, 
%
Tseng \emph{et al.}~\cite{tseng2020cross} first  pre-trains a model on the source domain and then learns feature-wise transformation on four target domains to adapt target distribution.
Guo \emph{et al.}~\cite{guo2020broader} reveal that advanced few-shot meta-learning methods underperform simple finetuning approach under the cross-domain few-shot setting. 
ConFT~\cite{das2021importance} constructs negative pairs using unlabeled source samples and target samples to boost few-shot generalization in the finetune stage.
}
%
Li \emph{et al.}~\cite{li2022ranking} treats CD-FSC as an image retrieval task and minimizes task-irrelevant features while preserving more transferable discriminative information.
\wx{
Although these methods achieve competitive performance, they still suffer from overfitting and ignorant of prior learned transferable knowledge due to the lack of training samples in the finetuning phase.
On the other hand, \cite{chen2019catastrophic} reveals that
if only a limited number of training samples are available, overly maintaining pre-train knowledge would be harmful to the performance on target dataset and leads to negative transfer.
}
Therefore,  it's vital to pay more attention to the target-related knowledge (as shown in Figure~\ref{figure1}) when conducting knowledge preservation.

\wx{
The above observations inspired this work, we propose a novel Adaptive
Semantic Consistency (ASC) framework, which adaptively preserves target-related information in the source domain.
}
\wx{
Specifically, 
}
we design an adaptive weight assignment strategy that re-weight each reused source sample according to how similar it is to target domain. The similarity between each source sample and target domain is measured by their euclidean distance in the feature space. Subsequently, the value of the distance is transformed into the new weight of source samples.
\wx{
On top of the re-weighted samples, we introduce an extra auxiliary source model, which is a copy of pre-trained model, and its parameters are kept unchanged when finetuning. We further propose a semantic consistency regularization to regularize the features of source samples output by the source model and target model to be consistent during the finetuning phase. 
}
\wx{
%
In this way, ASC enables preserving the target-related prior knowledge from source domain and thus alleviates the overfitting problem.
We conduct extensive experiments on standard CD-FSC benchmarks, and the results indicate the effectiveness of the proposed ASC. 
}

\wx{In summary, } our contributions are as follows:
\begin{itemize}
\item 
\wx{
We propose a novel plug-and-play ASC framework for cross-domain few-shot classification, which incorporates source knowledge to alleviate overfitting limited target samples. 
}
\item 
\wx{
An adaptive weight assignment strategy and a semantic consistency regularization are further designed to aggregate target-related transferable information from source domain.
}
\item 
\wx{
Extensive experiments on multiple benchmarks verify the effectiveness of the proposed ASC, and ASC achieves competitive results compared to previous methods.
}
\end{itemize}

\section{Related work}
\textbf{Few-Shot Classification} FSC aims to classify samples of novel classes with an extremely small training set. Currently, the predominant method of FSC problem is meta-learning, which could be roughly divided into two categories, metric-based and optimization-based. The goal of metric-based methods \cite{snell2017prototypical, vinyals2016matching, sung2018learning}is “learning to compare”, that is, classifying a query sample based on its similarity to the labeled support set, and the optimization-based methods \cite{rusu2018meta, finn2017model, ravi2017optimization}aim to adapt a pre-trained generalizable model to novel tasks with few labeled training samples. Meta-learning has achieved great success on traditional FSC problem. However, \cite{chen2019closer, tian2020rethinking}suggest that simply finetuning a model pre-trained on the whole meta-training dataset could have competitive or even better performance than meta-learning methods. When it comes to CD-FSC setting, the performance gap between finetuning and meta-learning gets bigger.\\
\\
\textbf{Cross-Domain Few-Shot Classification} The difference between CD-FSC and traditional few-shot classification is that the source and target datasets are from different domains in CD-FSC but from the same domain under the traditional few-shot classification setting. \cite{tseng2020cross}firstly proposes a benchmark that pre-trains a model on a single source domain then finetunes it on four target domains. However, \cite{guo2020broader} spots its limitation that all of the target domains used in \cite{tseng2020cross} are natural images. To this end, another four target domains are added in \cite{guo2020broader}, which consist of natural images, satellite images, dermatology images, and radiology images respectively. There are two important issues for CD-FSC, one is the domain shift between source and target datasets. To solve this problem, some methods try to obtain a prior model with high generalization capability, aiming to alleviate the influence of domain shift. \cite{tseng2020cross}utilize feature-transformation layers to simulate different feature distributions in training stage. \cite{liang2021boosting}uses an autoencoder to reconstruct the images and utilize the reconstructed images as noise input.
Another widely existing issue for CD-FSC is the overfitting problem due to the lack of support samples. The model is prone to overfitting to limited support samples and fails to generalize to query set after finetuning. To alleviate this, \cite{das2021importance}reuses unlabeled data from source domain to serve as negative samples in a modified supervised contrastive loss, which can improve few-shot generalization. \cite{zhengcross} argues that a number of weakly activated elements would serve as noise and lead to overfitting so they propose a feature denoising operation to remove the close-to-zero elements when finetuning. In this work, we also focus on the overfitting but alleviate it in another way. We notice that the model is under a risk of ignoring prior learned transferable knowledge due to the lack of support samples so we attempt to preserve them when finetuning.\\
\begin{figure*}[htbp]
\centering
  \includegraphics[width=\linewidth]{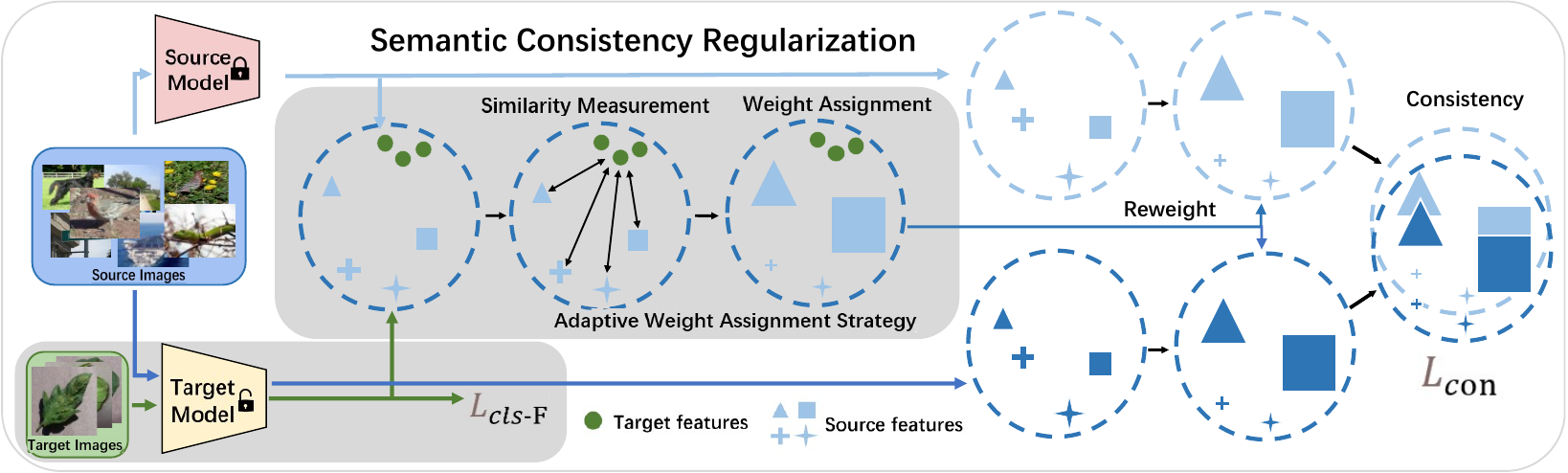}
  \caption{Overview of our Adaptive Semantic Consistency (ASC) framework. Firstly, we measure the similarity between each source sample and target domain by their euclidean distance and reweight the source images based on the source-target similarity. After that, we compute the consistency loss $L_{con}$ by measuring the distance between the output of source and target model on the same images and jointly minimize $L_{con}$ with classification loss $L_{cls-F}$.}
  \label{figure2}
\end{figure*}
\\
\textbf{Transfer Learning} The goal of transfer learning is to transfer the knowledge learned from a source task to a target task \cite{pan2010survey}. One of the most commonly used methods, finetuning, is to improve the performance of a pre-trained model on downstream tasks by adjusting all the parameters or the last few layers. In this paper, we set all the parameters of the pre-trained model trainable. To preserve prior learned knowledge and alleviate overfitting, a common way is to add regularization when finetuning. Some regularization-based methods for transfer learning have good performances. For example, \cite{xuhong2018explicit}constrains the effective search space around the initial solution to avoid overfitting, and \cite{li2019delta}regularizes the output of selected layers to be consistent, aiming to address the problem of \cite{xuhong2018explicit}where the constraint is too weak or too strong. However, both of them are under a setting where the size of target dataset is large. When there are only few training samples in target dataset, these methods would lead to negative transfer and meet a  catastrophic forgetting $\&$ negative transfer dilemma, illustrated in \cite{chen2019catastrophic}. To this end, we reuse images from source domain and propose an adaptive weight assignment strategy to highlight the source samples with high similarity to target domain in order to aggregate informative target-related knowledge from source
domain.

\section{Methodology}
\subsection{Problem formulation}
We start by formulating the CD-FSC problem. There are two available datasets, a large-scale source dataset and a small-scale target dataset, and they are from different domains. We use $\mathcal{D}_s=\{(x_s^i, y_s^i) | x_s^i \in X_s, y_s^i \in Y_s, i = 1, 2,..., M\}$ to denote the source dataset where M is the total number of the source samples, and $\mathcal{D}_t = (X_t, Y_t)$ to denote the target dataset. Their label sets have no overlap, $i.e., Y_s \cap Y_t = \emptyset$. $\mathcal{D}_t$ could be further divided into a labeled support set $\mathcal{D}_{supp} = \{(x_{supp}^i, y_{supp}^i)| x_{supp}^i \in X_t, y_{supp}^i \in Y_t, i = 1, 2, ..., N \times K\}$ and an unlabeled query set $\mathcal{D}_q = \{x_q^i|x_q^i \in X_t\}$, and these two sets share the same labels. This is also known as “N-way K-shot” few-shot learning, as there are N classes in $\mathcal{D}_{supp}$ and each class has K samples.

\subsection{Overview}
\label{3_2}
Since transfer learning can outperform meta-learning under CD-FSC setting, we choose transfer learning as our pipeline. In the pre-training phase, a model is trained from scratch on source dataset $\mathcal D_s$ with a classification loss. Then, the pre-trained model is made in two copies. One is named source model with fixed parameters and the other is named target model with tunable parameters. After that, target model would be finetuned on support set $\mathcal{D}_{supp}$ meanwhile regularized by source model and would be eventually evaluated on query set $\mathcal{D}_q$. 
Our proposed adaptive semantic consistency module would not change the finetuning procedure of the inserted CD-FSC methods. Concretely, we simply introduce an extra consistency loss ${L}_{con}$, which has a completely separate computation procedure from the existing classification loss ${L}_{cls-F}$ of the CD-FSC methods. In each epoch of the finetuning phase, we sample a batch of unlabeled source images and simultaneously feed them into the aforementioned two models. ${L}_{con}$ is computed by the two models' output and is jointly minimized with ${L}_{cls-F}$. After finetuning, source model is discarded and only target model is used for inference. An overview of the finetuning procedure with our plug-and-play module is depicted in Fig. \ref{figure2}.

\begin{table*}
\resizebox{\linewidth}{!}{
\begin{tabular}{cccccccccc}
\toprule
\multirow{2}{*}{Methods} & \multicolumn{8}{c}{5-way 1-shot} \\
\cmidrule{2-10}
 & ISIC & EuroSAT & CropDisease & ChestX & CUB & Cars & Places & Plantae & Ave.\\
\midrule
Finetuning$^{\dag}$ & 31.14{\small $\pm$0.57} & 52.43{\small $\pm$0.80} & 55.69{\small $\pm$0.92} & 21.62{\small $\pm$0.35} & 38.44{\small $\pm$0.70} & 30.88{\small $\pm$0.58} & 41.76{\small $\pm$0.75} & 35.25{\small $\pm$0.64} & 38.40\\
Finetuning + ASC (Ours) & \textbf{34.45}{\small $\pm$0.60} & 52.34{\small $\pm$0.88} & \textbf{66.68}{\small $\pm$0.89} & \textbf{22.77}{\small $\pm$0.38} & \textbf{43.83}{\small $\pm$0.75} & \textbf{34.51}{\small $\pm$0.65} & 41.71{\small $\pm$0.73} & \textbf{38.84}{\small $\pm$0.73} & \textbf{41.89}\\
\cmidrule{1-10}
ConFT & 34.47{\small $\pm$0.60}$^{\ddag}$ & 64.79{\small $\pm$0.80}$^{\ddag}$ & 69.71{\small $\pm$0.90}$^{\ddag}$ & 23.31{\small $\pm$0.40}$^{\ddag}$ & 45.57{\small $\pm$0.76} & 39.11{\small $\pm$0.76} & \textbf{49.97}{\small $\pm$0.86} & \textbf{43.09}{\small $\pm$0.78} & 46.25\\
ConFT + ASC (Ours) & \textbf{35.16}{\small $\pm$0.61} & \textbf{65.27}{\small $\pm$0.80} & \textbf{69.96}{\small $\pm$0.84} & 23.32{\small $\pm$0.43} & \textbf{46.47}{\small $\pm$0.78} & \textbf{39.26}{\small $\pm$0.75} & 49.80{\small $\pm$0.76} & 41.58{\small $\pm$0.76} & \textbf{46.35}\\
\cmidrule{1-10}
Supcon (Ours) & 35.81{\small $\pm$0.63} & 63.65{\small $\pm$0.85} & \textbf{71.55}{\small $\pm$0.83} & 22.86{\small $\pm$0.41} & 42.80{\small $\pm$0.75} & \textbf{37.21}{\small $\pm$0.71} & \textbf{49.16}{\small $\pm$0.78} & 39.60{\small $\pm$0.71} & 45.33\\
Supcon + ASC (Ours) & \textbf{36.08}{\small $\pm$0.64} & \textbf{64.95}{\small $\pm$0.79} & 71.42{\small $\pm$0.82} & \textbf{23.26}{\small $\pm$0.43} & \textbf{45.46}{\small $\pm$0.77} & 36.98{\small $\pm$0.68} & 48.54{\small $\pm$0.82} & \textbf{40.98}{\small $\pm$0.74} & \textbf{45.96}\\

\bottomrule
\end{tabular}
}
\resizebox{\linewidth}{!}{
\begin{tabular}{cccccccccc}
\toprule
\multirow{2}{*}{Methods} & \multicolumn{8}{c}{5-way 5-shot} \\
\cmidrule{2-10}
 & ISIC & EuroSAT & CropDisease & ChestX & CUB & Cars & Places & Plantae & Ave.\\
\midrule
Finetuning$^{\dag}$ & 47.78{\small $\pm$0.59} & 77.97{\small $\pm$0.65} & 87.44{\small $\pm$0.66} & 25.71{\small $\pm$0.40} & 65.06{\small $\pm$0.68} & 49.99{\small $\pm$0.76} & 70.30{\small $\pm$0.68} & 55.93{\small $\pm$0.68} & 60.02\\
Finetuning + ASC (Ours) & \textbf{50.76}{\small $\pm$0.62} & \textbf{80.00}{\small $\pm$0.61} & \textbf{89.80}{\small $\pm$0.54} & \textbf{26.70}{\small $\pm$0.45} & \textbf{67.48}{\small $\pm$0.70} & \textbf{53.72}{\small $\pm$0.70} & \textbf{70.76}{\small $\pm$0.64} & \textbf{59.46}{\small $\pm$0.70} & \textbf{62.34}\\
\cmidrule{1-10}
ConFT & 50.79{\small $\pm$0.60}$^{\ddag}$ & 81.52{\small $\pm$0.60}$^{\ddag}$ & 90.90{\small $\pm$0.60}$^{\ddag}$ & 27.50{\small $\pm$0.50}$^{\ddag}$ & 70.53{\small $\pm$0.75} & 61.53{\small $\pm$0.75} & 72.09{\small $\pm$0.68} & \textbf{62.54}{\small $\pm$0.76} & 64.68\\
ConFT + ASC (Ours) & \textbf{51.42}{\small $\pm$0.68} & \textbf{82.43}{\small $\pm$0.58} & \textbf{91.44}{\small $\pm$0.45} & \textbf{28.32}{\small $\pm$0.44} & \textbf{72.86}{\small $\pm$0.66} & \textbf{62.18}{\small $\pm$0.79} & 72.05{\small $\pm$0.69} & 62.20{\small $\pm$0.69} & \textbf{65.36}\\
\cmidrule{1-10}
Supcon (Ours) & 51.11{\small $\pm$0.68} & 82.50{\small $\pm$0.58} & 92.01{\small $\pm$0.46} & 28.03{\small $\pm$0.46} & 67.47{\small $\pm$0.73} & 58.27{\small $\pm$0.78} & 70.96{\small $\pm$0.72} & 61.00{\small $\pm$0.75} & 63.91\\
Supcon + ASC (Ours) & \textbf{53.60}{\small $\pm$0.65} & \textbf{84.22}{\small $\pm$0.57} & \textbf{92.86}{\small $\pm$0.41} & \textbf{28.57}{\small $\pm$0.47} & \textbf{72.09}{\small $\pm$0.65} & \textbf{62.18}{\small $\pm$0.72} & \textbf{71.61}{\small $\pm$0.71} & \textbf{63.32}{\small $\pm$0.67} & \textbf{66.08}\\
\bottomrule
\end{tabular}
}

\caption{5-way 1-shot/5-shot tasks few-shot classification accuracy(\%) with our ASC module integrated into Finetuning, ConFT, and our proposed Supcon. $^{\dag}$ denotes our reproduced results using the official released code from \cite{guo2020broader}. $^{\ddag}$ indicates the result reported in \cite{li2022ranking}.}
\label{table1}
\end{table*}

\subsection{Adaptive Weight Assignment Strategy}
\label{awas}
We propose an adaptive weight assignment strategy to highlight the source samples with high similarity to target domain and measure the similarity by the euclidean distance between source and target features. Firstly, we feed the support images $x_{supp}^i$ into the target model $f_t$ to obtain their features. Then, we average all the features of support images to get their prototype $z_{supp}$, which is used to represent the target domain. Meanwhile, we feed the source images $x_s^i$ into source model $f_s$ to gain their features. Subsequently, we can evaluate the similarity of each source image to target domain by the euclidean distance $d_{st}^i$ between its corresponding features and the prototype of support images. $z_{supp}$ and $d_{st}^i$ can be defined as:
\[
z_{supp} = \frac{1}{NK}\sum_{i=1}^{NK} f_t(x_{supp}^i) \tag{1}
\]
\[
d_{st}^i = d(f_s(x_s^i), z_{supp}) \tag{2}
\]
where $d(a,b)$ denotes the euclidean distance between the two elements $a$ and $b$. With the distance matrix $d_{st}$ that evaluates the similarity between target domain and all the source samples in a batch, the weight $w_i$ for the $i^{th}$ source samples can be defined as:
\[
w_i = (softmax(-d_{st}) \times B)_i  \tag{3} \label{weight}
\]
Here, we use ($\cdot$)$_i$ to denote the $i^{th}$ element in the brackets, and B, the batch size of unlabeled source samples, is multiplied to keep the magnitude unchanged.

\subsection{Semantic Consistency Regularization}
With the reweighted source samples, we further propose a semantic consistency regularization to constrain the features of source samples output by the aforementioned two models throughout the finetuning phase. In our ASC framework, the function of source model is to provide consistent semantic features of each source sample, which subsequently serve as a reference for target model. For the $i^{th}$ source image sampled in one finetuning epoch, we simultaneously feed it into the source and target models and compute the euclidean distance $d_{ss}^i$ of the two models' outputs. After that, we reweight each $d_{ss}^i$ with the weight adaptively computed in formulation \ref{weight}, and the consistency loss ${L}_{con}$ is the average of all the reweighted $d_{ss}^i$. We denote source and target models and the $i^{th}$ source sample as $f_s$, $f_t$, and $x_s^i$ respectively. Then, the consistency loss ${L}_{con}$ can be defined as:
 \[
{L}_{con} = \frac{1}{B}\sum_{i=1}^{B} w_i d_{ss}^i = \frac{1}{B}\sum_{i=1}^{B}w_i\Vert f_s(x_s^i) - f_t(x_s^i) \Vert_2 ^ 2 \tag{4} \label{lcon}
\]

On top of this, the total loss function of our ASC framework is formulated as:
\[
{L}_{ASC} = {L}_{cls-F} + \lambda{L}_{con} \tag{5} \label{lssr}
\]
where ${L}_{cls-F}$ is the aforementioned classification loss function of a specific CD-FSC method and $\lambda$ is a hyper-parameter that controls the weight of the consistency loss ${L}_{con}$.

\begin{table*}[htbp]
  \begin{tabular}{cccccccccc}
\toprule
methods& Hyperparameters & ISIC & EuroSAT & CropDisease & ChestX & CUB & Cars & Places & Plantae\\
\midrule
Supcon & learning rate & 5e-3 & 1e-3 & 5e-3 & 5e-3 & 5e-3 & 5e-3 & 5e-4 & 5e-3\\
ConFT \& Supcon + ASC & ${\tau}^*$ & 0.05 & 0.05 & 0.05 & 0.05 & 0.1 & 0.1 & 0.05 & 0.1\\
\midrule
Finetune + ASC & learning rate & 0.05 & 0.01 & 0.05 & 0.05 & 0.05 & 0.05 & 0.01 & 0.05\\
\bottomrule
\end{tabular}

\caption{Hyper-parameter details for different methods on each target dataset. * denotes the hyper-parameters that only appear in ConFT and Supcon.}
  \label{hyperpara}
\end{table*}

\subsection{Choices of CD-FSC Methods}
\label{choices_Lcls}
Since our ASC is a plug-and-play method, we can have multiple choices for CD-FSC methods to be inserted. The first one is Finetune \cite{guo2020broader}, which uses standard cross-entropy loss as classification loss $L_{cls-F}$. In this method, a linear classifier is added on the top of target model. We denote the linear classifier and cross-entropy loss as $W$ and $L_{CE}$ respectively, and the classification loss can be defined as:
\[
L_{cls-F} = L_{CE}(W(f_t(x_{supp})), y_{supp}) \tag{6}
\]
The second CD-FSC method we choose is ConFT \cite{das2021importance}, which also reuses unlabeled source samples (named as distractor) in finetuning phase. Different from us, the distractors in ConFT are utilized to construct negative pairs with each support sample in a modified supervised contrastive loss. With a distractor set $\mathcal I_{dt}$ and a subset $\mathcal I_{supp}$ of support set $\mathcal{D}_{supp}$, ConFT first construct an anchor-positive index set $P(i)$ and an anchor-negative index set $N(i)$ for each sample $x_{supp}^i$ in $\mathcal I_{supp}$. With the features of $x_{supp}^i$, positive sample, negative sample, and distractor denoted as $z_{supp}^i$, $z_p$, $z_n$ and $z_d$ respectively, the classification loss of ConFT can be defined as:
\[
L_{cls-F} = -\frac{1}{\left | I_{supp} \right |}\sum_{i \in I_{supp}}\frac{1}{\left | P(i) \right |}\sum_{p \in P(i)} log(l_{ip}),
\]
\[
l_{ip} = \frac{exp(\frac{z_i \cdot z_p}{\tau})}{exp(\frac{z_i \cdot z_p}{\tau}) + \sum_{n \in N(i)}exp(\frac{z_i \cdot z_n}{\tau}) + \sum_{d \in I_{dt}}exp(\frac{z_i \cdot z_d}{\tau})}  \tag{7}
\]
where $\tau$ is a temperature hyper-parameter.\\
Except for the two aforementioned methods, we also find that using a supervised contrastive loss (Supcon) \cite{khosla2020supervised} as classification loss in finetuning phase can also achieve competitive results. Therefore, we also insert our ASC module into Supcon to conduct experiments. We denote the features of the $i^{th}$ sample in $\mathcal{D}_{supp}$ as $z_i$. The classificaton loss in our proposed Supcon is defined as:
\[
L_{cls-F} = -\frac{1}{NK}\sum_{i=1}^{NK}\frac{1}{\left | P(i) \right |}\sum_{p \in P(i)}log\frac{exp(\frac{z_i \cdot z_p}{\tau})}{\sum_{j=1}^{NK}exp(\frac{z_i \cdot z_j}{\tau})} \tag{8}
\]
Here, $P(i) = \{ p \in \mathcal D_{supp} | p \neq i, y_{supp}^p = y_{supp}^i \}$ is the set of indices of all support samples that have the same label as $x_{supp}^i$.

\section{Experiments}
In this section, we evaluate our Adaptive Semantic Consistency framework on various target domains. We use the three CD-FSC methods mentioned in \ref{choices_Lcls} as baselines and integrate our ASC module into them, then we compare the results with and without our ASC module.

\begin{table}
  \begin{tabular}{ccc}
\toprule
\multirow{2}{*}{Datasets} & \multicolumn{2}{c}{5-way 5-shot} \\
\cmidrule{2-3}
 & w/o weight & with weight \\
\midrule
EuroSAT & 83.76 & \textbf{84.22} \\
CropDisease & 92.85 & 92.86 \\
CUB & 71.39 & \textbf{72.09} \\
Cars & 61.56 & \textbf{62.18} \\
\bottomrule
\end{tabular}
  \begin{tabular}{ccc}
\toprule
\multirow{2}{*}{Datasets} & \multicolumn{2}{c}{5-way 1-shot} \\
\cmidrule{2-3}
 & w/o weight & with weight  \\
\midrule
EuroSAT & 63.23 & \textbf{64.95} \\
CropDisease & 71.30 & \textbf{71.42} \\
CUB & 44.88 & \textbf{45.46} \\
Cars & 36.97 & 36.98 \\
\bottomrule
\end{tabular}
\caption{Results of ablation experiments of Supcon + ASC under 5-way 5/1-shot settings. "w/o weight" and "with weight" are the experiments without and with adaptive weight assignment strategy respectively.}
  \label{w_wo_weight}
\end{table}

\subsection{Experiments Settings}
\textbf{Datasets} Following the benchmarks \cite{tseng2020cross, guo2020broader}, we use mini-Imagenet as source dataset. It is a subset of ILSVRC-2012 \cite{russakovsky2015imagenet} and it contains 100 categories with 600 images in each category, and the first 64 categories are used for pre-training. There are 8 target datasets for finetuning. Among them, CUB \cite{welinder2010caltech}, Cars \cite{krause20133d}, Places \cite{zhou2017places} and Plantae \cite{van2018inaturalist} are proposed in \cite{tseng2020cross}. They consist of images from natural domains and have high similarity to mini-Imagenet. The other four datasets proposed in \cite{guo2020broader}, ISIC \cite{codella2019skin}, EuroSAT \cite{helber2019eurosat}, CropDisease \cite{mohanty2016using}, and ChestX \cite{wang2017chestx}, consists of images with different visual characteristics and they have low similarity to mini-Imagenet.\\
\textbf{Implementation details} For all the experiments, we use Resnet-10 \cite{he2016deep} as backbone, and we apply simple data-augmentation including random crop, random flip, and color jitter. As for the optimizer, we use SGD for Finetune \cite{guo2020broader}, and Adam \cite{kingma2014adam} for ConFT \cite{das2021importance} and our proposed Supcon. There are several hyper-parameters in our experiments, including the learning rate, batch size B of the source samples in formulation \eqref{lcon}, $\lambda$ in formulation \eqref{lssr}, and $\tau$ which only appears in ConFT \cite{das2021importance} and Supcon. Among them, B and $\lambda$ are fixed at 64 and 1 for all the experiments. There are some slight changes in learning rate and $\tau$ after a specific CD-FSC method is combined with ASC. The details of learning rate and $\tau$ are shown in table \ref{hyperpara}. \\
\textbf{Evaluation metric} The two previous methods and our proposed Supcon share the same pre-training stage, that is to train a randomly initialized model on the first 64 categories of mini-Imagenet with standard cross-entropy loss for 400 epochs. After that, we randomly select 5 classes in target dataset to finetune for 100 epochs. For each class, we randomly sample 5/1 images to make up support set and 15 images to make up query set. We report the mean classification accuracy over 600 few-shot learning tasks as well as 95\% confidence interval on the query set.

\begin{figure}[tbp]
	\centering
 
        \subfigure{
		\begin{minipage}[t]{0.475\linewidth}
			\centering
			\includegraphics[width=1\linewidth]{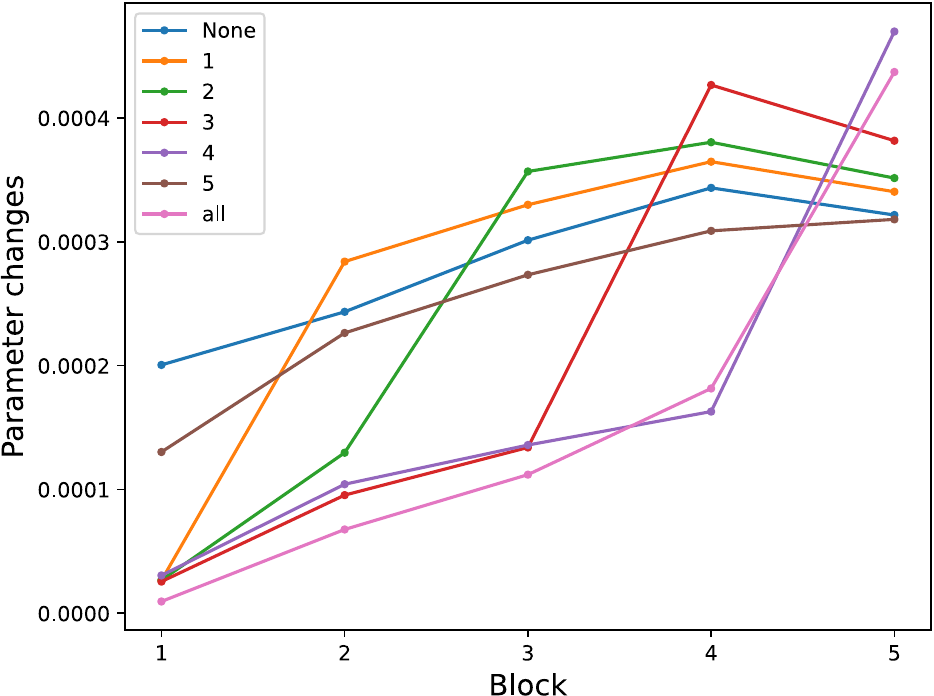}
        \caption*{EuroSAT}
		\end{minipage}
        }
        \subfigure{
        \begin{minipage}[t]{0.475\linewidth}
			\centering
			\includegraphics[width=1\linewidth]{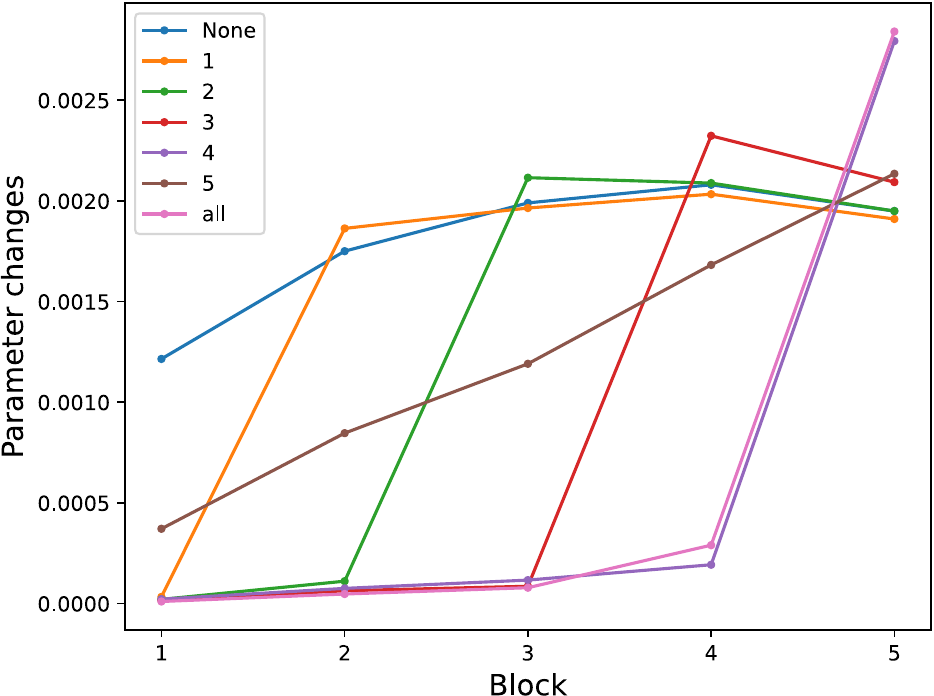}
        \caption*{CropDisease}
		\end{minipage}
	}
        \subfigure{
		\begin{minipage}[t]{0.475\linewidth}
			\centering
			\includegraphics[width=1\linewidth]{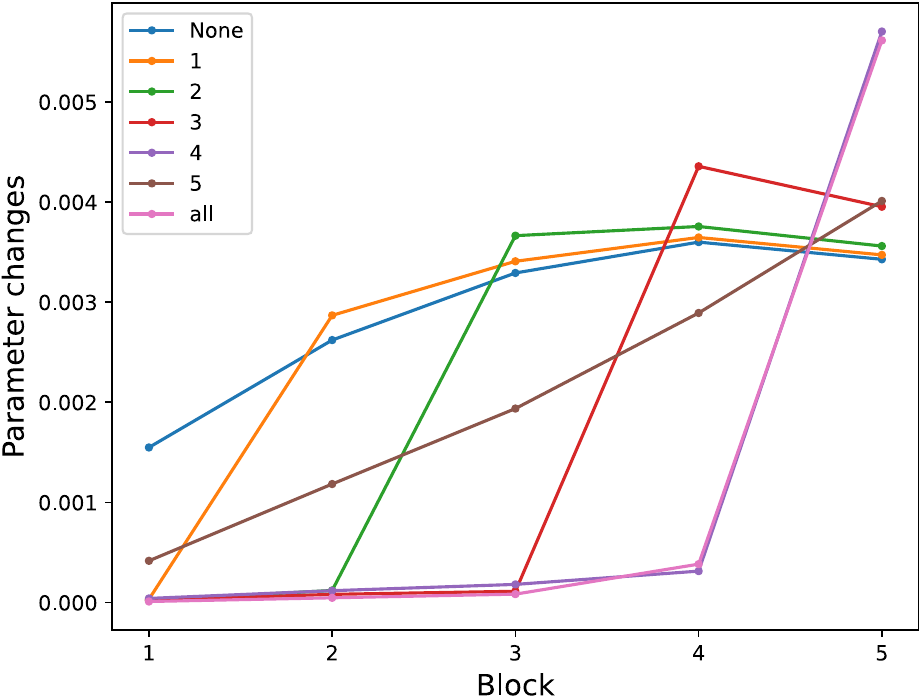}
        \caption*{CUB}
		\end{minipage}
        }
        \subfigure{
        \begin{minipage}[t]{0.475\linewidth}
			\centering
			\includegraphics[width=1\linewidth]{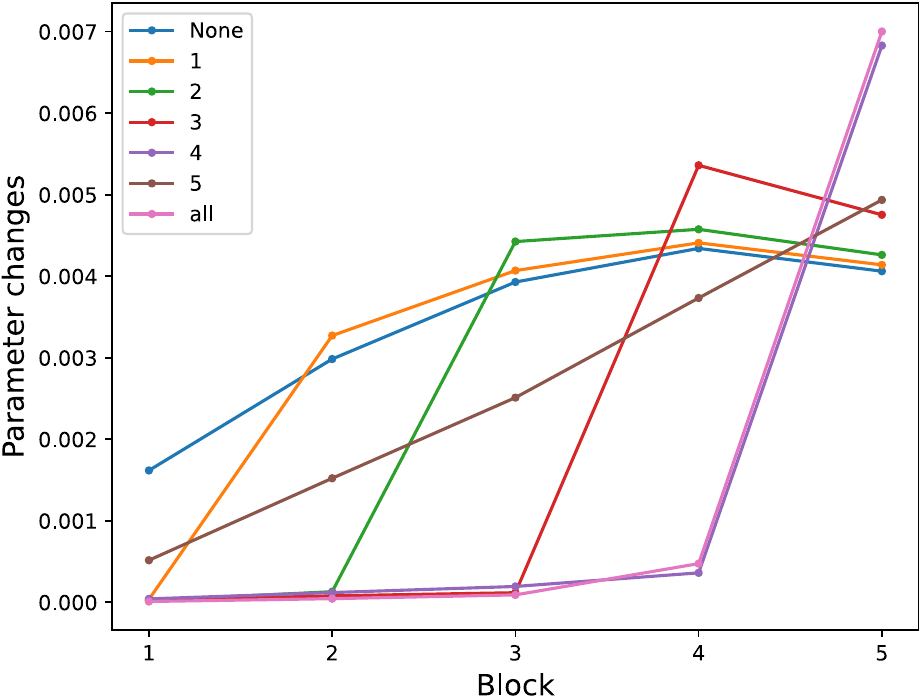}
        \caption*{Cars}
		\end{minipage}
	}
	\caption{Parameters change of each block after finetuning on 4 target datasets regularizing different levels of features output by different blocks of the backbone, which is denoted by different colors of lines in each subfigure.}
	\label{para_change}
\end{figure}

\subsection{Results Comparison} In our experiments, we integrate our adaptive semantic consistency module into Finetuning \cite{guo2020broader}, ConFT \cite{das2021importance} and our proposed Supcon to evaluate its effect on alleviating overfitting by preserving target-related prior knowledge. We can learn from the results shown in table \ref{table1} that our proposed ASC module provides an improvement in most cases.

\subsection{Ablation study}
In this subsection, we study the characteristics of our ASC framework. We pick EuroSAT, CropDisease, CUB and Cars to conduct ablation expriments. Among them, EuroSAT and CropDisease have low similarity to mini-Imagenet while CUB and Cars have high similarity to mini-Imagenet.\\

\textbf{Effect of Adaptive Weight Assignment Strategy.} We mentioned in section \ref{1} that the adaptive weight assignment strategy applied in our method has the capacity of highlighting prior learned target-related knowledge. To examine its effectiveness, we conduct ablation experiments with and without adaptive weight assignment strategy using Supcon loss as the classification loss under both 5/1-shot settings. According to the results shown in table \ref{w_wo_weight}, we can learn that our proposed adaptive weight assignment strategy can provide improvements in most cases.\\
\\
\textbf{Comparison between ASC and DELTA.} Similar to our method, DELTA  \cite{li2019delta} is also a regularization-based transfer learning method and it also proposes a fixed source model to assist finetuning. There are two differences between DELTA and our method. One is that our method uses source images for regularization while DELTA uses target images. The other is that our method regularizes the semantic-level features while DELTA aligns mid-level feature. We argue that DELTA would lead to negative transfer under CD-FSC setting. To examine our hypothesis, we follow the settings of DELTA to conduct experiments with Supcon loss as classification loss under 5-way 5-shot setting. 

\begin{table}[tbp]
\resizebox{\linewidth}{!}{
  \begin{tabular}{ccccc}
\toprule
 Regularized block & EuroSAT & CropDisease& CUB & Cars\\
\midrule
None (baseline) & 82.50 & 92.01 & 67.47 & 58.27 \\
1 & 83.79 & 92.29 & 68.24 & 59.10\\
2 & 83.91 & 92.40 & 68.62 & 59.48\\
3 & 84.12 & 92.48 & 68.71 & 59.35\\
4 & \textbf{84.31} & 92.31 & 68.66 & 59.04\\
5 & 84.22 & \textbf{92.86} & \textbf{72.09} & \textbf{62.18}\\
All & 84.27 & 92.49 & 71.34 & 61.49 \\
\bottomrule
\end{tabular}
}
\caption{5-way 5-shot results of Supcon + ASC with different layers' output is regularized.}
  \label{diff_layer}
\end{table}

\begin{figure}[tbp]
	\centering
 
        \subfigure{
		\begin{minipage}[t]{0.22\linewidth}
			\centering
			\includegraphics[width=1\linewidth]{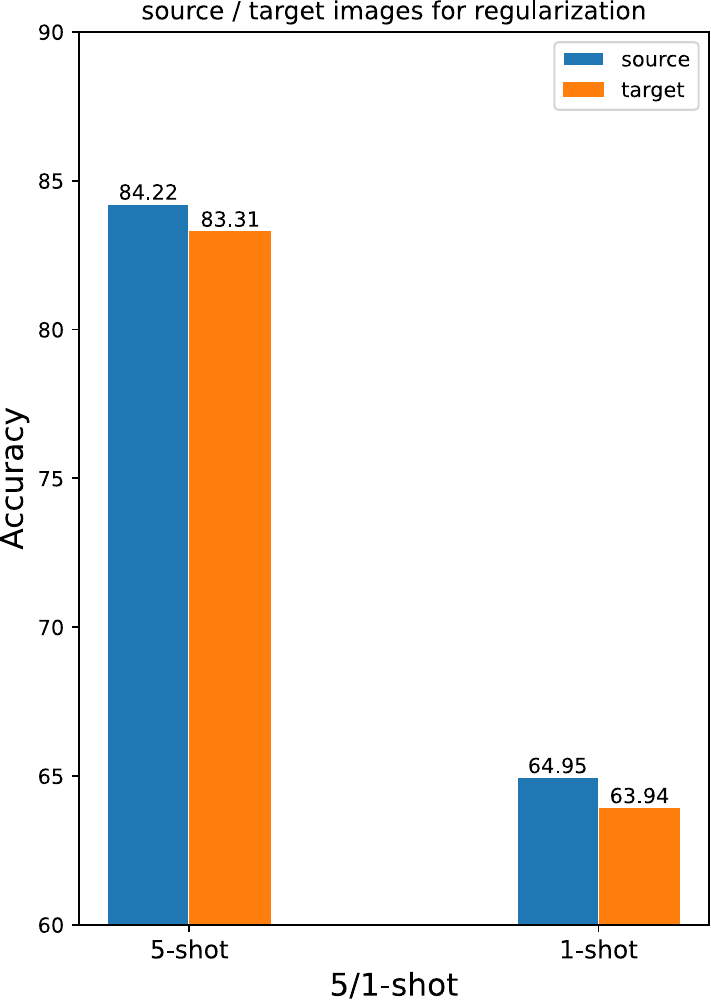}
        \caption*{EuroSAT}
		\end{minipage}
        }
        \subfigure{
        \begin{minipage}[t]{0.22\linewidth}
			\centering
			\includegraphics[width=1\linewidth]{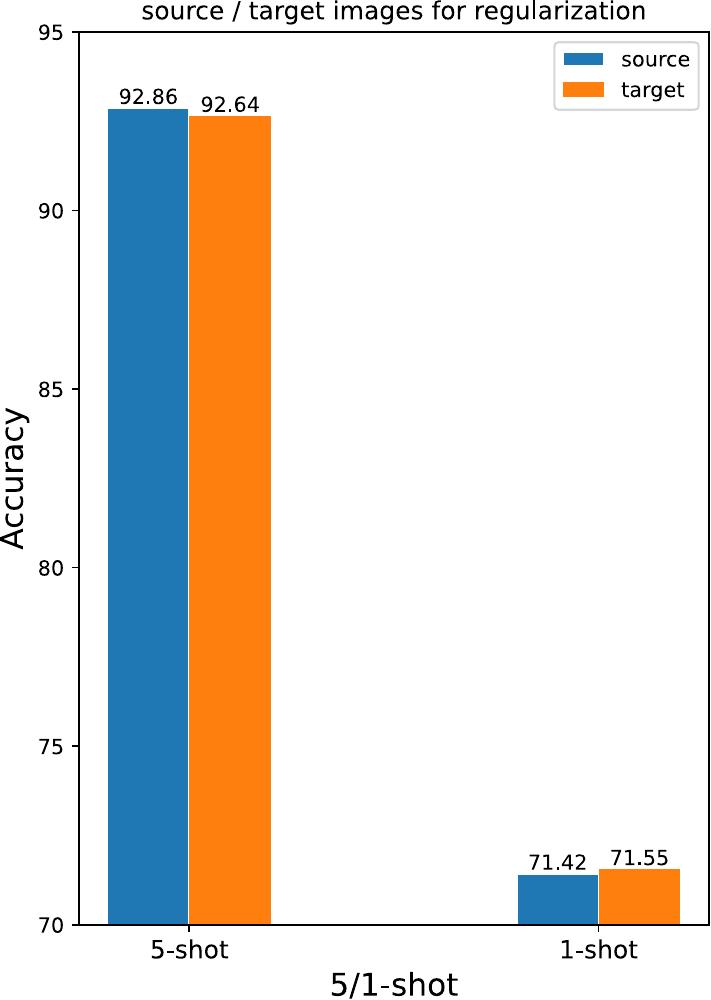}
        \caption*{CropDisease}
		\end{minipage}
	   }
        \subfigure{
		\begin{minipage}[t]{0.22\linewidth}
			\centering
			\includegraphics[width=1\linewidth]{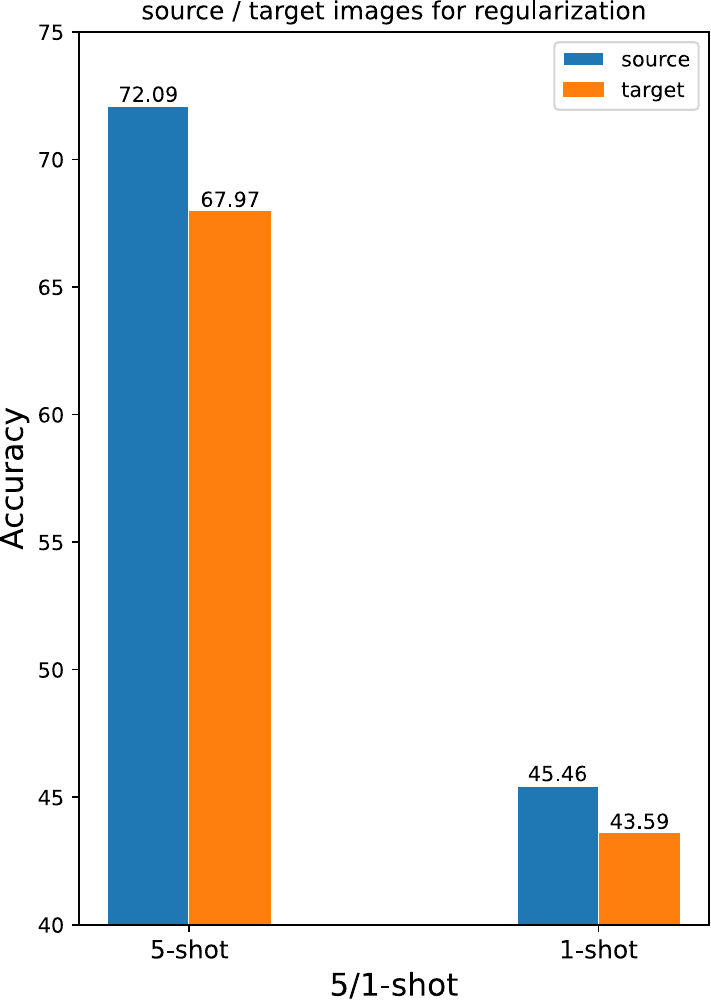}
        \caption*{CUB}
		\end{minipage}
        }
        \subfigure{
        \begin{minipage}[t]{0.22\linewidth}
			\centering
			\includegraphics[width=1\linewidth]{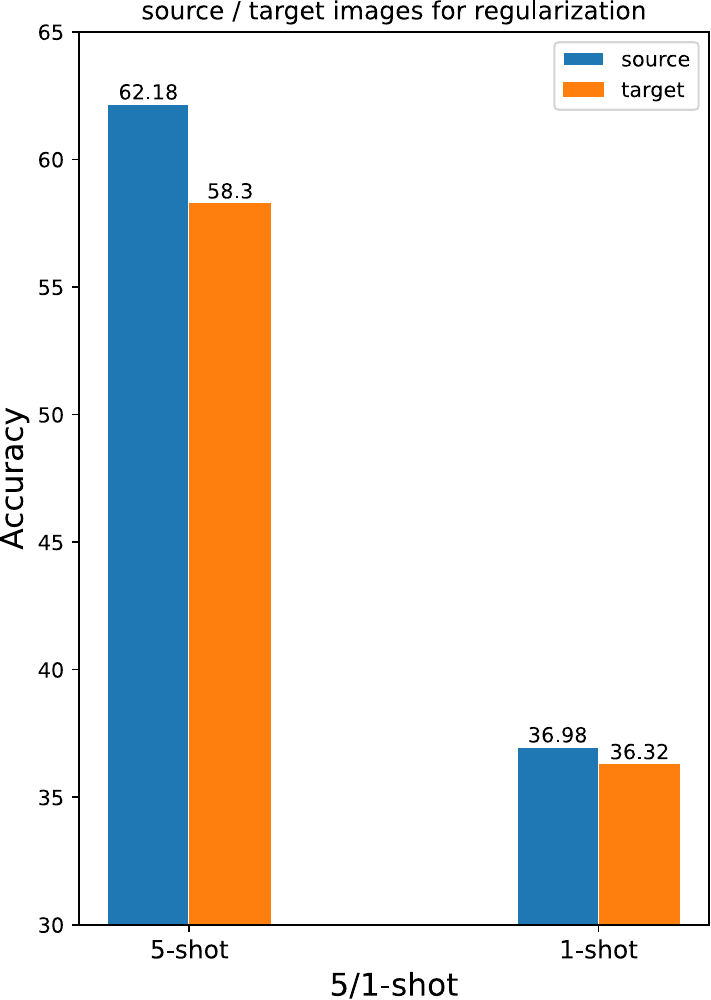}
        \caption*{Cars}
		\end{minipage}
	}
	\caption{5-way 5/1-shot results of Supcon + ASC with either source images or target images for semantic regularization.}
	\label{target_consistency}
\end{figure}

\begin{figure}[htbp]
\centering
\includegraphics[width=\linewidth]{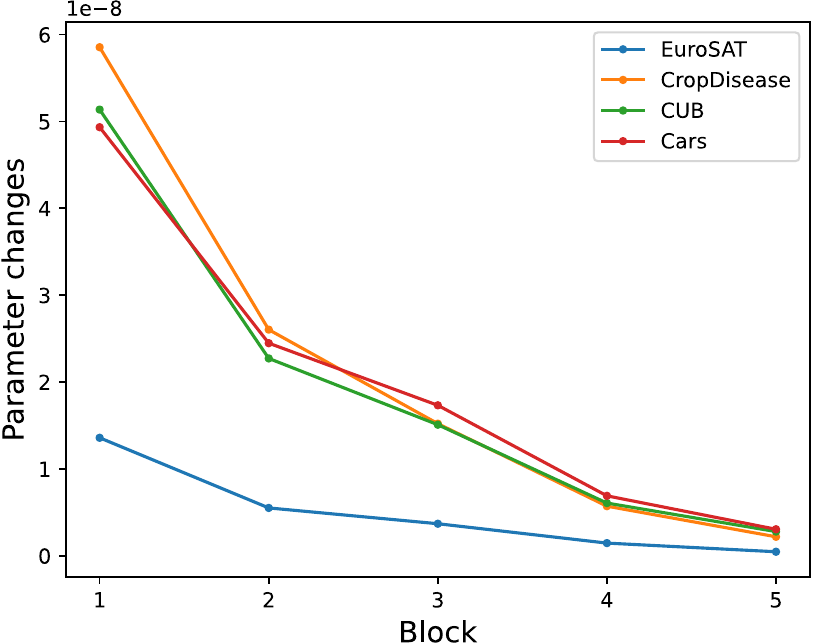}
\caption{Parameters change of each block after finetuning on 4 target datasets with target images to conduct semantic consistency regularization.}
\label{tgtreg_para}
\end{figure}

First, we imitate DELTA to regularize the mid-level features of source samples and compare the model's performance with different layers' output is regularized. As our backbone (Resnet-10) can be divided into 5 blocks, we separately impose regularization on each block's output and all of them in different experiments and show the results in table \ref{diff_layer}.
We can learn from table \ref{diff_layer} that in most circumstances, the target model's performance drops significantly if mid-level features instead of semantic-level features are regularized. To better understand the impact of mid-level / semantic-level features regularization, we further analyze how they influence the parameters of each block. We use $\theta$ and $\hat{\theta}$ to denote the parameters of pre-trained model and finetuned model respectively. The parameters changes can be defined as $(\hat{\theta} - \theta)^2$, averaged over all parameters in each block, on 100 tasks. The results of each block's parameters changes are shown in Figure~\ref{para_change}, and we can learn that the mid-level features regularization has a much stronger impact on constraining model's parameters than semantic-level features regularization. With the mid-level regularization, parameters of the blocks below the regularized block would hardly change, and we hypothesize that would result in negative transfer.\\
What's more, we also use target images to conduct semantic consistency regularization as DELTA does and the results are shown in Fig. \ref{target_consistency}. We can observe that target model's performance drops on all the 4 target datasets if we use target images instead of source images for regularization, especially on CUB and Cars. We argue that is because the consistency loss has a great impact on constraining the parameters of target model if the features of target images are regularized. As shown in Fig. \ref{tgtreg_para}, the the parameters of target model hardly change after finetuning on the 4 target datasets, which may results in excessive preservation of prior learned knowledge and further leads to negative transfer. \\

\begin{figure*}[htbp]
	\centering
	
        \subfigure{
        \begin{minipage}[t]{0.1\linewidth}
			\centering
			\includegraphics[width=1\linewidth, height=1\linewidth]{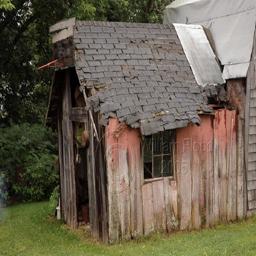}
		\end{minipage}
        \begin{minipage}[t]{0.1\linewidth}
			\centering
			\includegraphics[width=1\linewidth, height=1\linewidth]{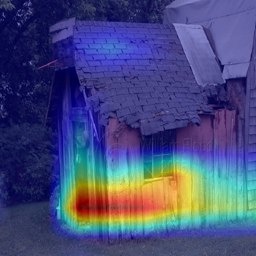}
		\end{minipage}
        \begin{minipage}[t]{0.1\linewidth}
			\centering
			\includegraphics[width=1\linewidth, height=1\linewidth]{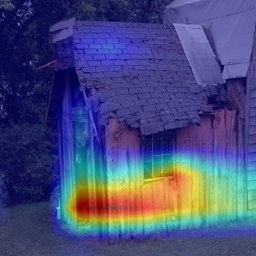}
		\end{minipage}
	}
	\subfigure{
		\begin{minipage}[t]{0.1\linewidth}
			\centering
			\includegraphics[width=1\linewidth, height=1\linewidth]{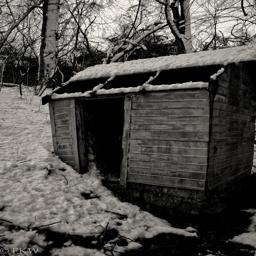}
		\end{minipage}
        \begin{minipage}[t]{0.1\linewidth}
			\centering
			\includegraphics[width=1\linewidth, height=1\linewidth]{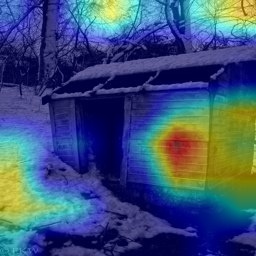}
		\end{minipage}
        \begin{minipage}[t]{0.1\linewidth}
			\centering
			\includegraphics[width=1\linewidth, height=1\linewidth]{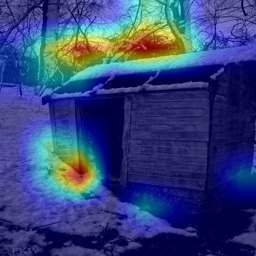}
		\end{minipage}
        \hspace{.1in}
        }
        \subfigure{
		\begin{minipage}[t]{0.1\linewidth}
			\centering
			\includegraphics[width=1\linewidth, height=1\linewidth]{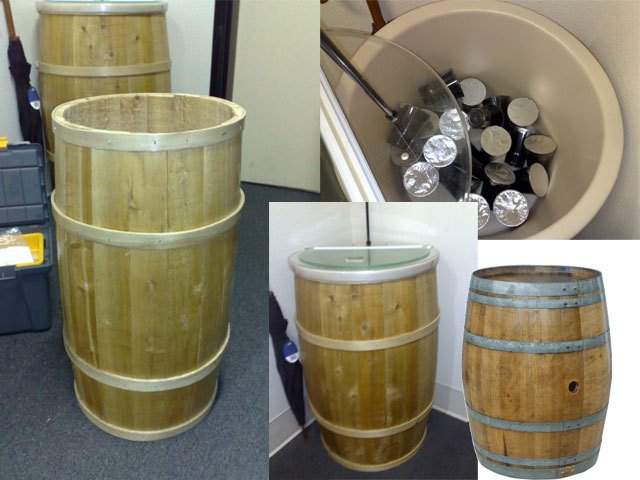}
		\end{minipage}
        \begin{minipage}[t]{0.1\linewidth}
			\centering
			\includegraphics[width=1\linewidth, height=1\linewidth]{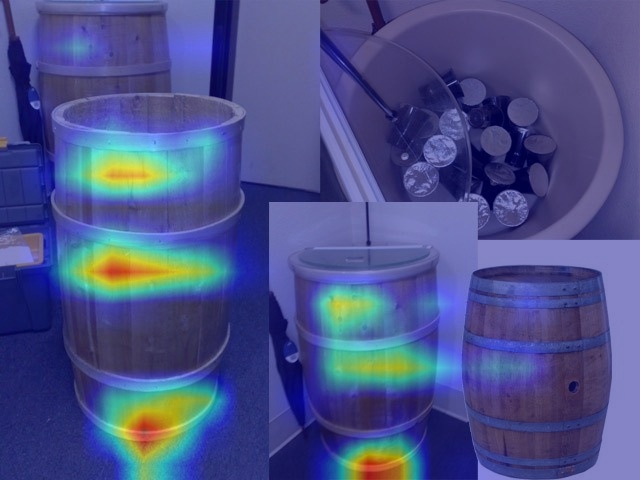}
		\end{minipage}
        
        }
        \setcounter{subfigure}{0}
    \subfigure[\large Places-Support]{
		\begin{minipage}[t]{0.1\linewidth}
			\centering
			\includegraphics[width=1\linewidth, height=1\linewidth]{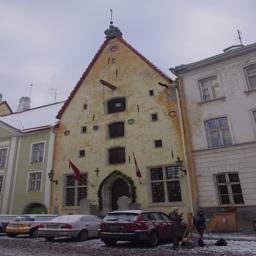}
		\end{minipage}
        \begin{minipage}[t]{0.1\linewidth}
			\centering
			\includegraphics[width=1\linewidth, height=1\linewidth]{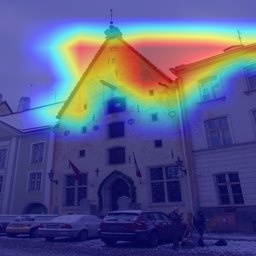}
        \caption*{ours}
		\end{minipage}
        \begin{minipage}[t]{0.1\linewidth}
			\centering
			\includegraphics[width=1\linewidth, height=1\linewidth]{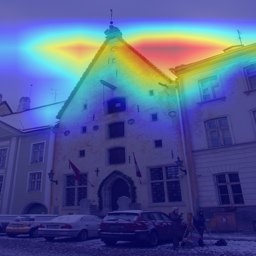}
        \caption*{baseline}
		\end{minipage}
        }
        \subfigure[\large Places-Query]{
		\begin{minipage}[t]{0.1\linewidth}
			\centering
			\includegraphics[width=1\linewidth, height=1\linewidth]{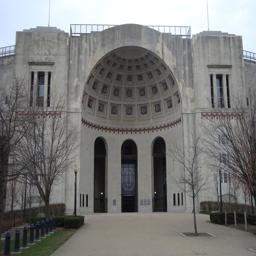}
		\end{minipage}
        \begin{minipage}[t]{0.1\linewidth}
			\centering
			\includegraphics[width=1\linewidth, height=1\linewidth]{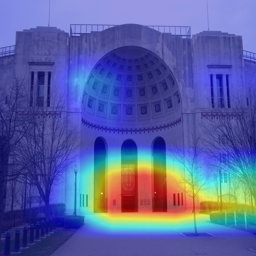}
        \caption*{ours}
		\end{minipage}
        \begin{minipage}[t]{0.1\linewidth}
			\centering
			\includegraphics[width=1\linewidth, height=1\linewidth]{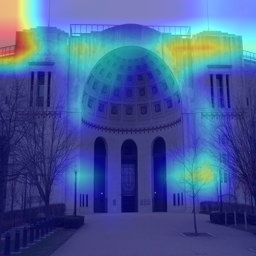}
        \caption*{baseline}
		\end{minipage}
        \hspace{.1in}
        }
        \subfigure[\large mini-Imagenet]{
        \begin{minipage}[t]{0.1\linewidth}
			\centering
			\includegraphics[width=1\linewidth, height=1\linewidth]{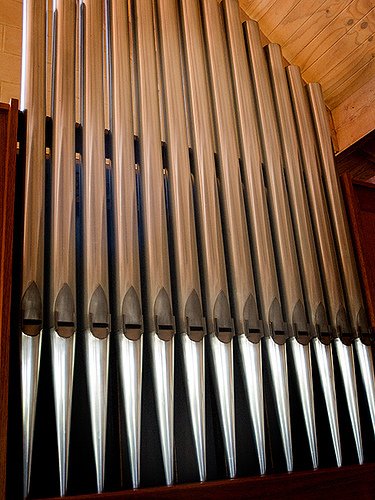}
		\end{minipage}
        \begin{minipage}[t]{0.1\linewidth}
			\centering
			\includegraphics[width=1\linewidth, height=1\linewidth]{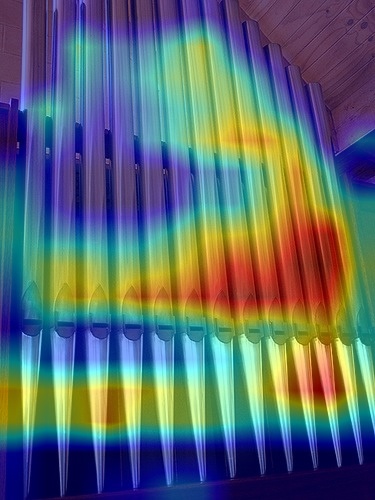}
        \caption*{Pre-train}
		\end{minipage}
        }

 
        \subfigure{
        \begin{minipage}[t]{0.1\linewidth}
			\centering
			\includegraphics[width=1\linewidth, height=1\linewidth]{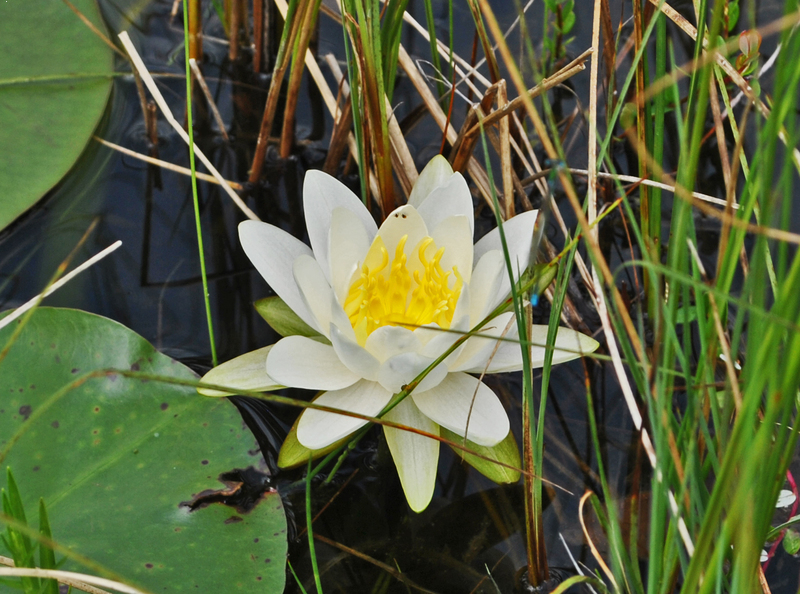}
		\end{minipage}
        \begin{minipage}[t]{0.1\linewidth}
			\centering
			\includegraphics[width=1\linewidth, height=1\linewidth]{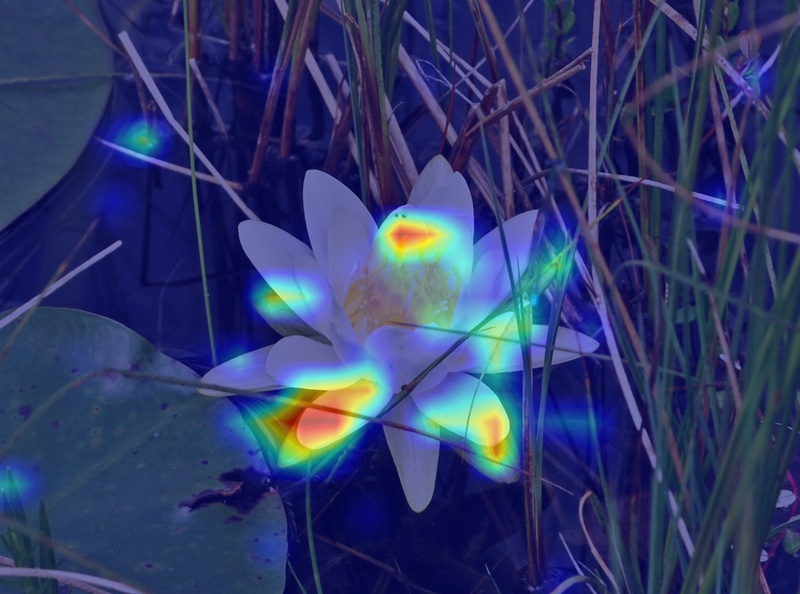}
		\end{minipage}
        \begin{minipage}[t]{0.1\linewidth}
			\centering
			\includegraphics[width=1\linewidth, height=1\linewidth]{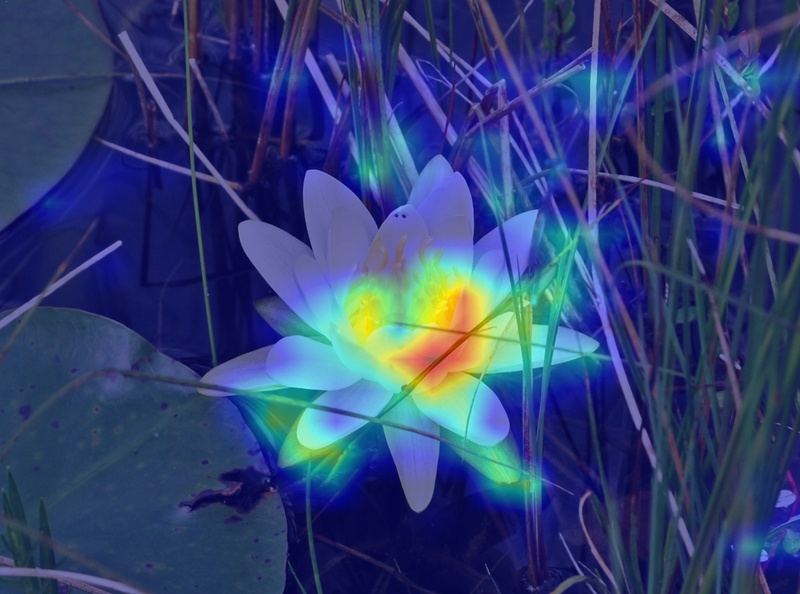}
		\end{minipage}
	   }
        \subfigure{
        \begin{minipage}[t]{0.1\linewidth}
			\centering
			\includegraphics[width=1\linewidth, height=1\linewidth]{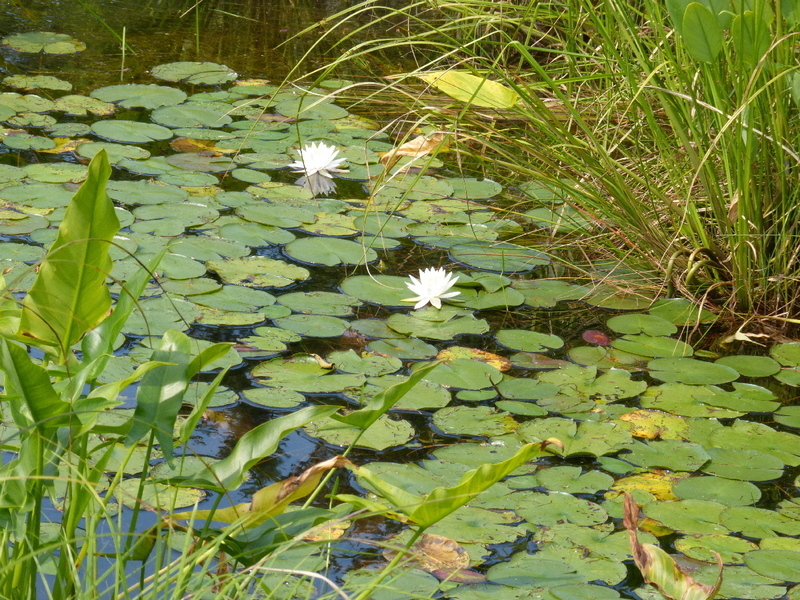}
		\end{minipage}
        \begin{minipage}[t]{0.1\linewidth}
			\centering
			\includegraphics[width=1\linewidth, height=1\linewidth]{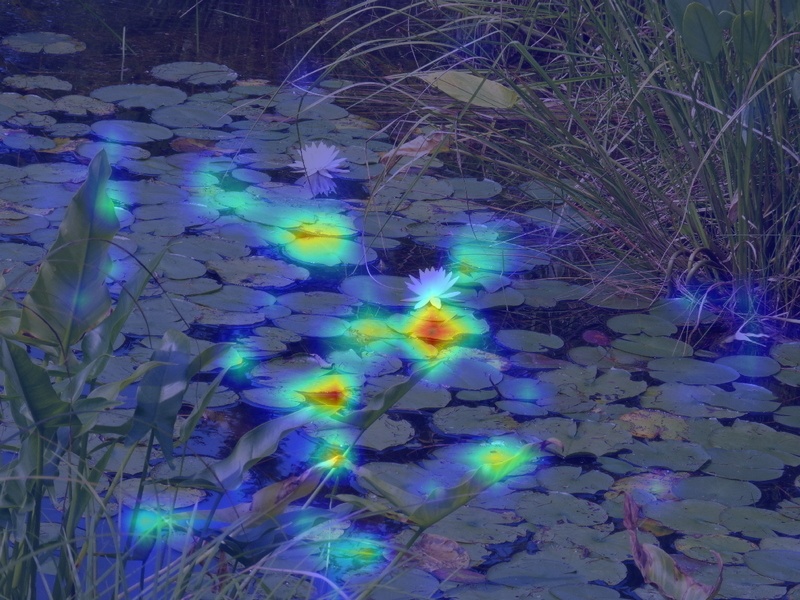}
		\end{minipage}
        \begin{minipage}[t]{0.1\linewidth}
			\centering
			\includegraphics[width=1\linewidth, height=1\linewidth]{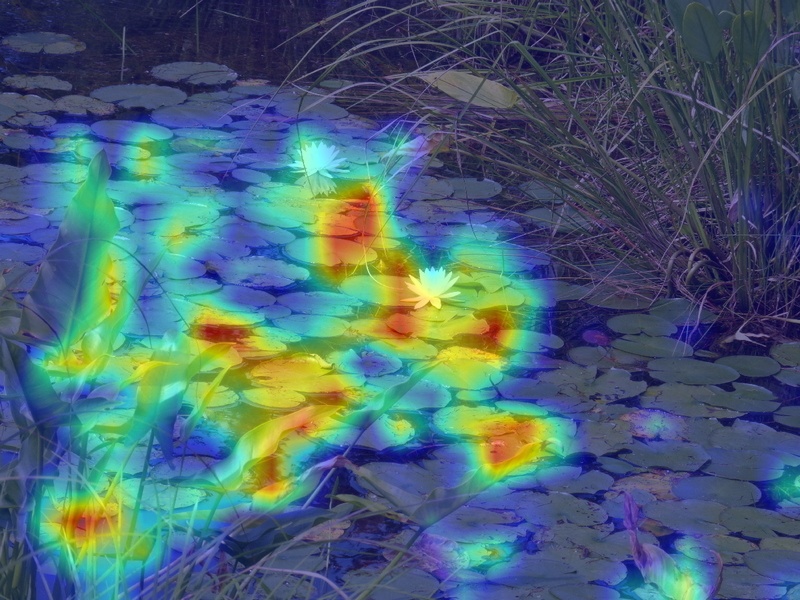}
		\end{minipage}
        \hspace{.1in}
	   }
                \subfigure{
		\begin{minipage}[t]{0.1\linewidth}
			\centering
			\includegraphics[width=1\linewidth, height=1\linewidth]{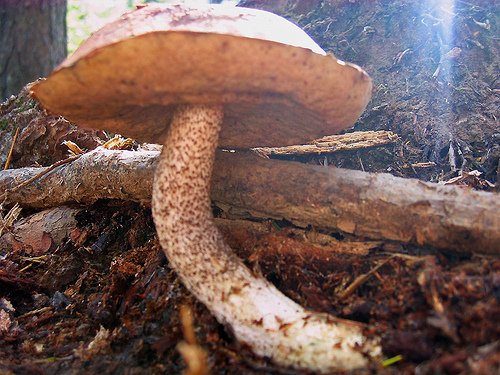}
		\end{minipage}
        \begin{minipage}[t]{0.1\linewidth}
			\centering
			\includegraphics[width=1\linewidth, height=1\linewidth]{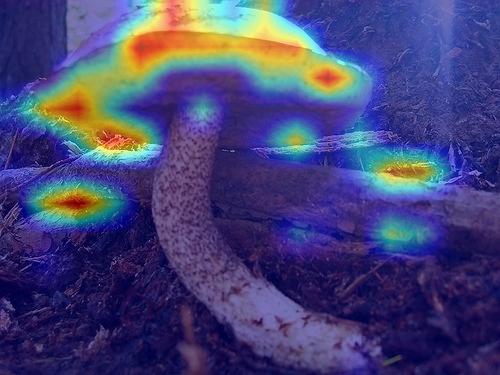}
		\end{minipage}
        
        }
        \setcounter{subfigure}{0}
        \subfigure[\large Plantae-Support]{
        \begin{minipage}[t]{0.1\linewidth}
			\centering
			\includegraphics[width=1\linewidth, height=1\linewidth]{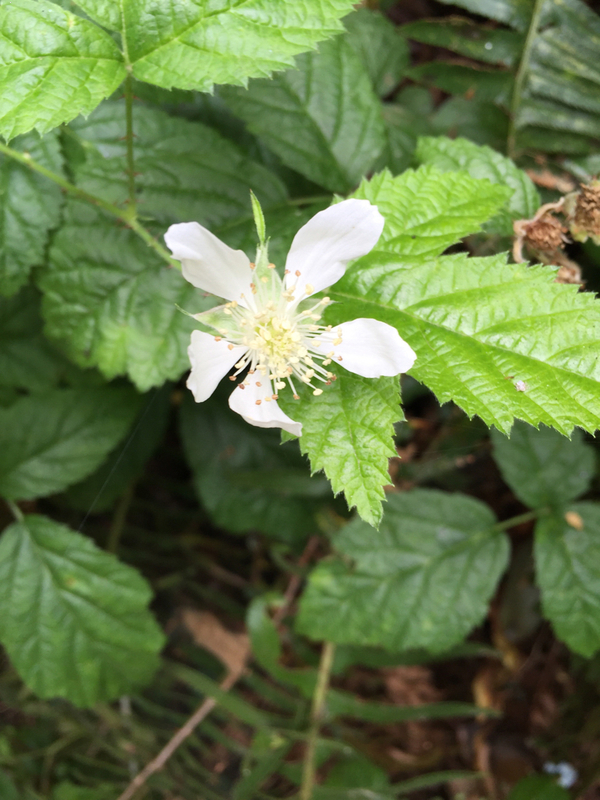}
		\end{minipage}
        \begin{minipage}[t]{0.1\linewidth}
			\centering
			\includegraphics[width=1\linewidth, height=1\linewidth]{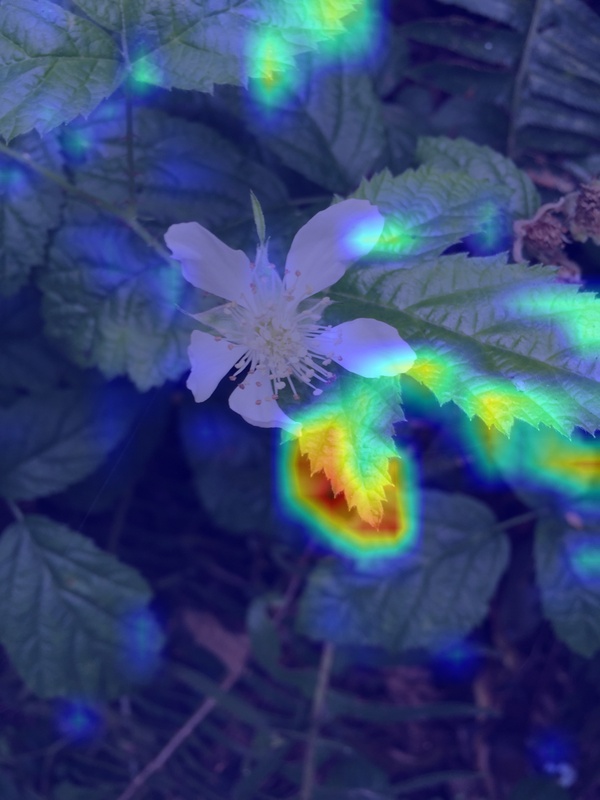}
        \caption*{ours}
		\end{minipage}
        \begin{minipage}[t]{0.1\linewidth}
			\centering
			\includegraphics[width=1\linewidth, height=1\linewidth]{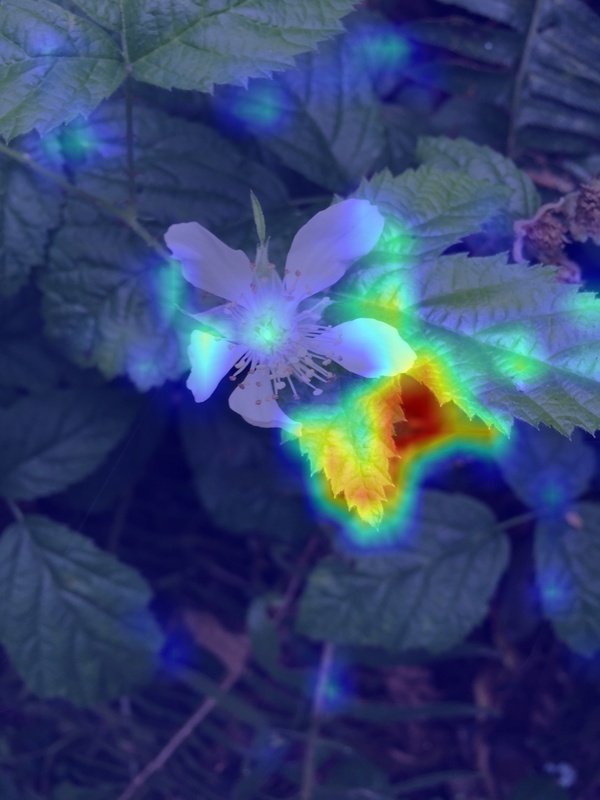}
        \caption*{baseline}
		\end{minipage}
	}
        \subfigure[\large Plantae-Query]{
        \begin{minipage}[t]{0.1\linewidth}
			\centering
			\includegraphics[width=1\linewidth, height=1\linewidth]{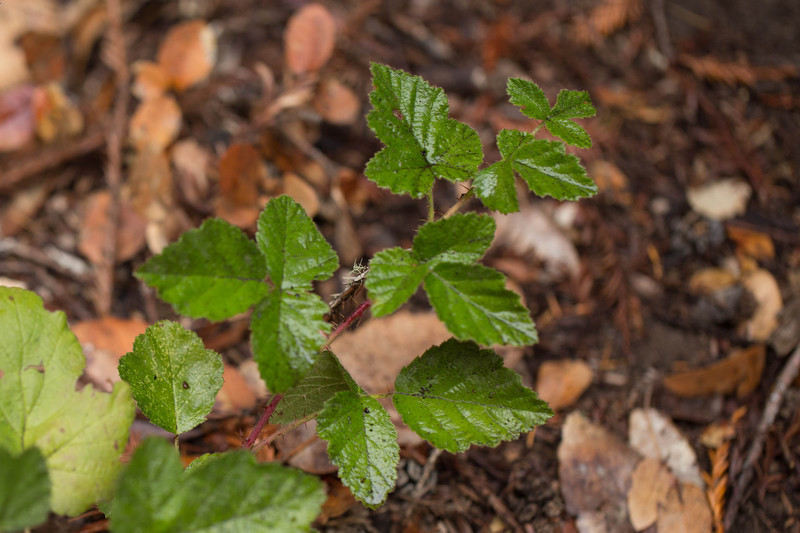}
		\end{minipage}
        \begin{minipage}[t]{0.1\linewidth}
			\centering
			\includegraphics[width=1\linewidth, height=1\linewidth]{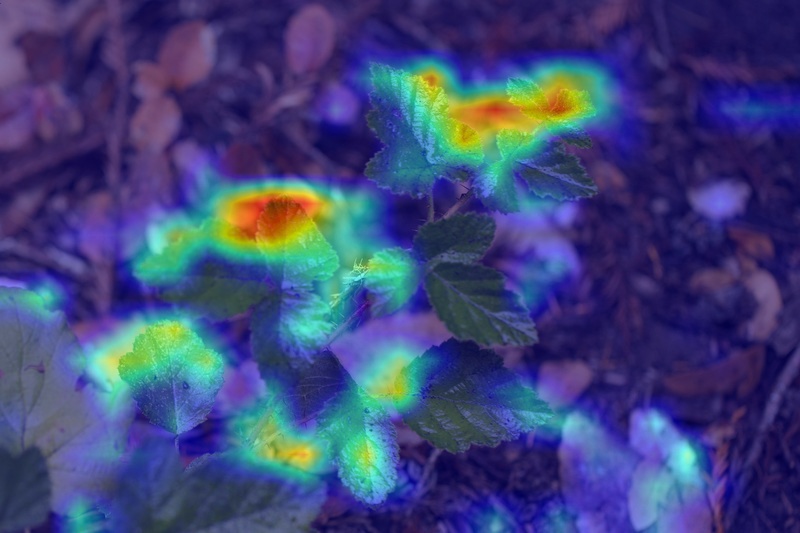}
        \caption*{ours}
		\end{minipage}
        \begin{minipage}[t]{0.1\linewidth}
			\centering
			\includegraphics[width=1\linewidth, height=1\linewidth]{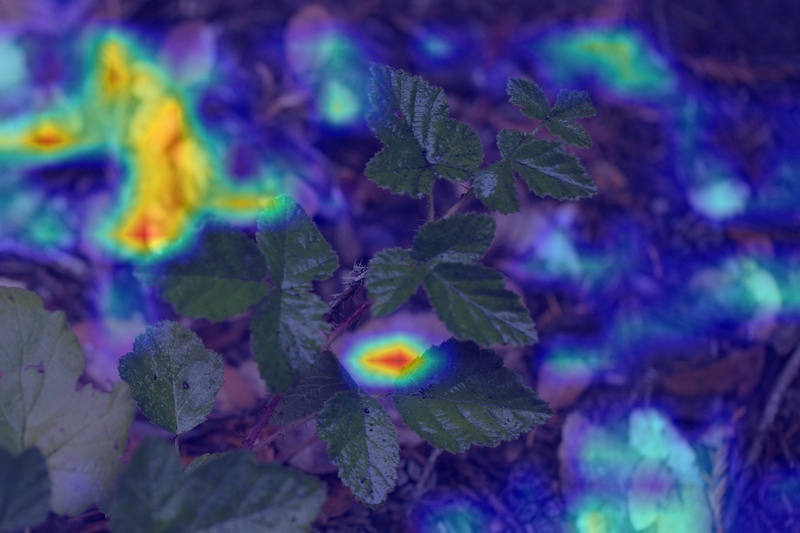}
        \caption*{baseline}
		\end{minipage}
        \hspace{.1in}
	   }
        \subfigure[\large mini-Imagenet]{
		\begin{minipage}[t]{0.1\linewidth}
			\centering
			\includegraphics[width=1\linewidth, height=1\linewidth]{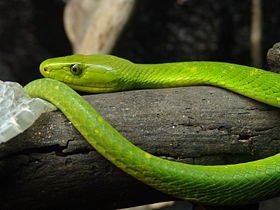}
		\end{minipage}
        \begin{minipage}[t]{0.1\linewidth}
			\centering
			\includegraphics[width=1\linewidth, height=1\linewidth]{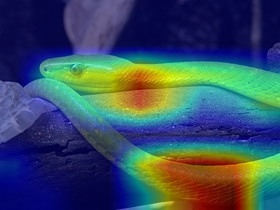}
        \caption*{Pre-train}
		\end{minipage}
        
        }
	\caption{Class-activation map of pre-trained model and two finetuned model with and without ASC module. For each target dataset, we choose two categories and for each category, we display one support image, one query image, and one corresponding source image with similar features.}
	\label{CAM}
\end{figure*}

\textbf{Batch size of source samples.} As the number of total finetuning epochs is 100 and the batch size of sampled source images in each epoch is 64, only $\frac{1}{6}$ of the source dataset is utilized in our method. Intuitively, the larger the batch size, the more source images are "visualized" and the more prior knowledge is preserved. To examine how the number of "visualized" source images influences our method's performance, we conduct experiments with different source images batch size ranging from 4 to 256 under 5-way 5-shot setting and use Supcon as classification loss. According to the experimental results shown in table \ref{table5}, we can learn that the batch size has a different impact on different target datasets. EuroSAT and CropDisease are robust to the batch size while CUB and Cars prefer large batch size. We hypothesize that is related to the similarity between source and target datasets. As CUB and Cars are more similar to mini-Imagenet than the other two, it's beneficial for them to aggregate more informative knowledge from source domain.\\

\begin{table}[htbp]
  \begin{tabular}{ccccc}
\toprule
batch size & EuroSAT & CropDisease & CUB & Cars\\
\midrule
4 & 84.16 & 92.71 & 71.42 & 61.24\\
8 & 84.12 & 92.62 & 71.70 & 61.48\\
16 & 84.16 & 92.77 & 71.85 & 61.86\\
32 & 84.15 & 92.74 & 72.03 & 62.02\\
64 & \textbf{84.22} & \textbf{92.86} & 72.09 & 62.18\\
128 & 84.14 & 92.76 & 72.18 & 62.37\\
256 & 84.13 & 92.81 & \textbf{72.24} & \textbf{62.43}\\
\bottomrule
\end{tabular}
\caption{Experimental results of Supcon + ASC with different source images batch size under 5-way 5-shot setting.}
  \label{table5}
\end{table}

\textbf{Top Similarity Regularization} As we mentioned above, it's of great importance to pay more attention to target-related knowledge when conducting knowledge preservation. Our proposed ASC module achieves this by introducing an adaptive weight assignment strategy. Furthermore, it's intuitive to ask that if we only sample the source images with top similarity to target domain, will it be beneficial to target tasks? The answer is yes. To examine this, we conduct further experiments and design a source images sample strategy to select the most similar source images. First, we feed all the images in source dataset and support set into the pre-trained model to obtain their features. Then, we compute the prototype of the whole support set and each category in source dataset. The similarity between each source category and support set is measured by the euclidean distance between their prototype. We sort the source categories by their similarity to support set from largest to smallest and only the top m categories are used to conduct semantic consistency regularization. We choose the value of m ranging from 4 to 24 to conduct experiments on Supcon + ASC under 5-way 5-shot setting using Supcon loss as classification loss on CUB, Cars, Places, and Plantae. We can observe from the results shown in Table~\ref{topm} that only using the source images with high similarity to target domain instead of the whole source dataset can improve the performance of our ASC module. Besides, we also notice that if the similarity between source and target datasets is small, it's beneficial to apply a small value of m. As the similarity to mini-Imagenet of these four target datasets is Places, CUB, Cars, Plantae in descending order, the optimal choice of the value of m for these four datasets is also in descending order. However, the limitation of this source images selection strategy is that it has to utilize the labels of source images, which are not available in our setting.

\begin{table}[t]
{
  \begin{tabular}{ccccc}
\toprule
Value of m & CUB & Cars & Places & Plantae\\
\midrule
4 & 72.32 & 62.59 & 71.51 & \textbf{64.62} \\
8 & 72.40 & \textbf{62.62} & 71.58 & 63.68\\
12 & 72.49 & 62.57 & 72.22 & 63.52\\
16 & \textbf{72.54} & 62.35 & \textbf{72.70} & 63.60\\
24 & 72.13 & 62.24 & 71.46 & 62.92\\
64 & 72.09 & 62.18 & 71.61 & 63.32\\
\bottomrule
\end{tabular}
}
\caption{Experimental results of Supcon + ASC under 5-way 5-shot setting using source images with top similarity to target domain to conduct semantic consistency regularization.}
  \label{topm}
\end{table}

\subsection{Grad-CAM Visualization}
To evaluate the effect of our ASC module on alleviating overfitting and preserving target-related transferable knowledge, we visualize the class-activation maps of "Ours", "Baseline" and "Pre-train" using grad-CAM \cite{selvaraju2017grad}. "Pre-train" refers to pre-trained model without finetuning. "Baseline" refers to the target model finetuned on Places and Plantae with Supcon loss as classification loss without our ASC module, "Ours" denotes the model finetuned on the two aforementioned target datasets using Supcon loss and our ASC module. For each target dataset, we select two categories and choose one support image, and one query image from each category. What's more, we also select one source image for each target category which contains target-related knowledge. The results are shown in Fig. \ref{CAM}.\\
The finetuned model can extract discriminative features in support images. For example, the texture of shed, the edge of inclined roof, the petal of lotus and jagged edges of leaf. However, it suffers from overfitting and fails to perform well on some query images with features that do not appear or are not so obvious in support images. For example, the texture of the query shed is much more smooth than that in support shed and is ignored by finetuned model; the query house has a plain roof instead of a inclined roof; the size of lotus is too small and the lotus leaves take up the main part of the query image; the edge of query leaf is much smaller than the support leaf. These features in query images appear in source images but are ignored when finetuning. For instance, the smooth texture of query shed also appears on wooden barrel; the plain-roof building and pipe organ both contain pillars; the edge of lotus leaves and mushrooms looks similar and the stripe of snake's skin is also small like the query leaf's edge. Our ASC module can alleviate this problem by aggregating informative target-related knowledge from source domain. We can observe from Fig \ref{CAM} that the model trained with our ASC module can still focus on discriminative features that also exist in source images.

\section{Conclusion}
We propose a novel plug-and-play framework to alleviate overfitting in cross-domain few-shot classification. We apply an adaptive weight assignment strategy and a semantic consistency regularization in finetuning phase to aggregate prior learned target-related knowledge. Experimental results on eight target datasets show that our proposed ASC module provides an improvement on existing transfer-learning based CD-FSC methods.

\bibliographystyle{ACM-Reference-Format}
\bibliography{ref}


\end{document}